\documentclass{article}

\PassOptionsToPackage{numbers, compress}{natbib}

\usepackage[preprint]{neurips_2023}

\usepackage[utf8]{inputenc} 
\usepackage[T1]{fontenc}    
\usepackage{hyperref}       
\usepackage{url}            
\usepackage{booktabs}       
\usepackage{amsfonts}       
\usepackage{nicefrac}       
\usepackage{microtype}      
\usepackage[dvipsnames]{xcolor}        
\usepackage{xspace}
\usepackage{subfig, graphicx}
\usepackage{multicol, multirow}
\usepackage{array, wrapfig}
\usepackage[numbers, compress]{natbib}

\usepackage{graphicx}
\usepackage{pythonhighlight}

\newcolumntype{P}[1]{>{\centering\arraybackslash}p{#1}}

\title{Formatting Instructions For NeurIPS 2023}

\author{%
  Bin Xiao$^1$ \quad Lujun Gui$^2$ \\ 
  \quad
  \textbf{Lei Su$^1$ \quad Weipeng Chen$^1$ } \\
  $^1$Baichuan Inc. \quad $^2$ Beijing Institute of Technology \\
   $\{$xiaobin, sulei, chenweipeng$\}$@baichuan-inc.com\\ 
   $\{$lujun.gui$\}$@bit.edu.cn\\
   $$https://github.com/XiaoBin1992/clover$$ \\ 
}
  
\newcommand{\Sys}{\texttt{Clover-2}\xspace}
\newcommand{\clover}{\texttt{Clover}\xspace}

\begin{document}

\title{\Sys: Accurate Inference for Regressive Lightweight Speculative Decoding}

\date{}
\maketitle

\begin{abstract}

Large Language Models (LLMs) frequently suffer from inefficiencies, largely attributable to the discord between the requirements of auto-regressive decoding and the architecture of contemporary GPUs.
Recently, regressive lightweight speculative decoding has garnered attention for its notable efficiency improvements in text generation tasks.
This approach utilizes a lightweight regressive draft model, like a Recurrent Neural Network (RNN) or a single transformer decoder layer, leveraging sequential information to iteratively predict potential tokens.
Specifically, RNN draft models are computationally economical but tend to deliver lower accuracy, while attention decoder layer models exhibit the opposite traits.
This paper presents \Sys, an advanced iteration of \clover, an RNN-based draft model designed to achieve comparable accuracy to that of attention decoder layer models while maintaining minimal computational overhead.
\Sys enhances the model architecture and incorporates knowledge distillation to increase Clover's accuracy and improve overall efficiency.
We conducted experiments using the open-source Vicuna 7B and LLaMA3-Instruct 8B models.
The results demonstrate that \Sys surpasses existing methods across various model architectures, showcasing its efficacy and robustness.

\end{abstract}

\section{Introduction}
\label{sec:intro}

Generative Large Language Models (LLMs)~\cite{gpt2,chatgpt,gpt3}, exemplified by models such as GPT, have significantly transformed the field of artificial intelligence. These models showcase exceptional adaptability, extending their applications from creative writing to engaging in human-like chatbot conversations. Their profound understanding of natural language has enhanced human-computer interactions by automating tasks that require contextual sensitivity.
Nonetheless, LLMs encounter efficiency challenges when deployed on GPUs, primarily due to their sequential text generation mechanism, which involves two distinct phases: prefilling and decoding. The prefilling phase processes the entire input sequence to produce the initial token, whereas the decoding phase generates subsequent tokens iteratively, leveraging the input and previously generated tokens. The decoding phase, characterized by its repeated small-batch token processing cycles, leads to suboptimal utilization of GPU resources. This inefficiency in the decoding process represents a significant bottleneck in leveraging the full potential of these high-capacity models.

Speculative decoding \cite{pmlr-v202-leviathan23a, chen2023accelerating} is an acceleration technique devised to address the performance constraints associated with sequential text generation. This approach enhances computational efficiency by generating multiple tokens per step, while maintaining output consistency. The technique involves employing one or more lightweight draft models to predict several subsequent tokens with minimal computational overhead. These preliminary token predictions are then verified by the target model, allowing for the consolidation of token generations within a single iteration.

\begin{figure}[th]
\centering
\includegraphics[width=0.75\linewidth]{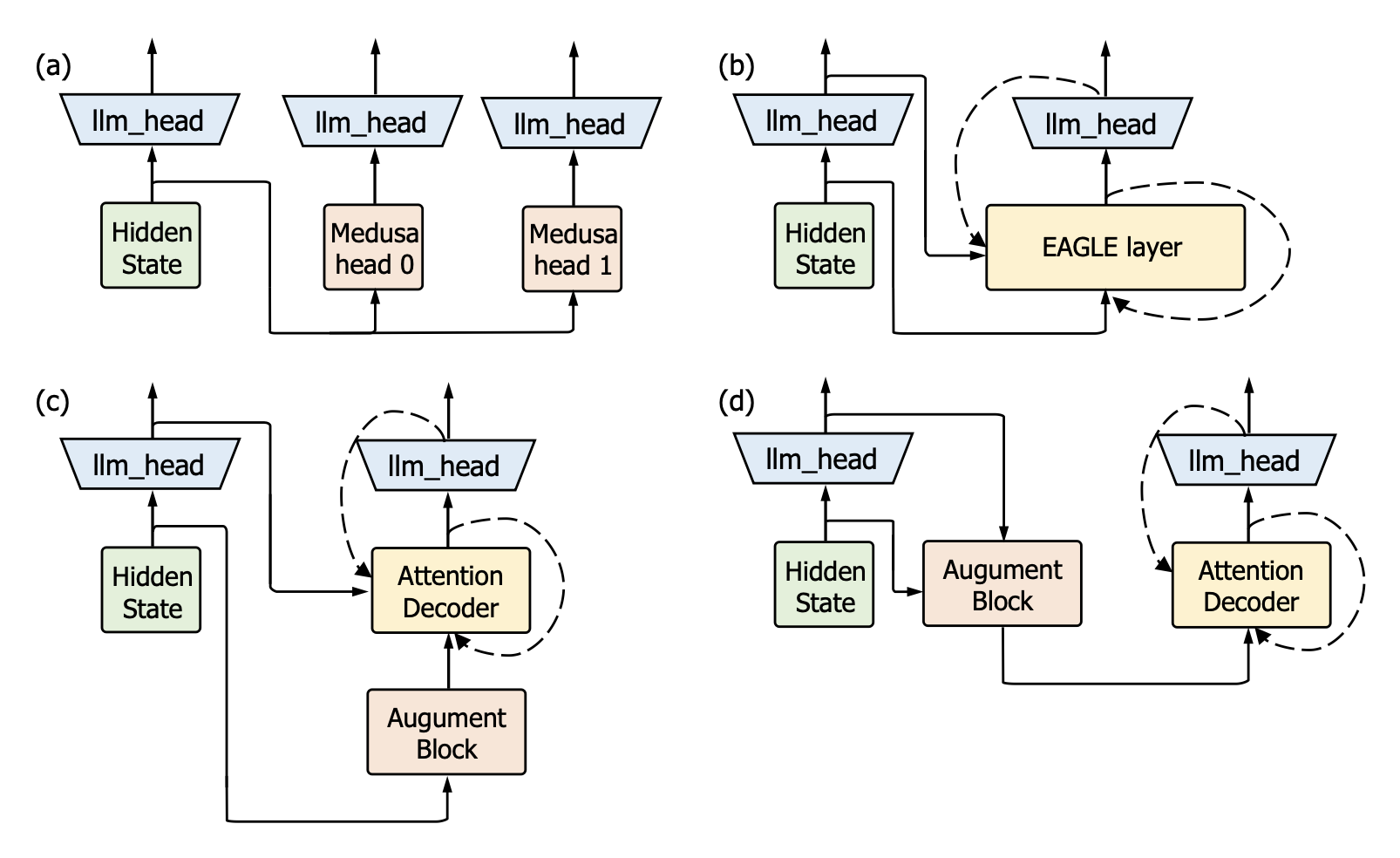}
\caption{Overview of (a) Medusa, (b) EAGLE, (c) \clover, and (d) \Sys.}
\label{fig:intro}
\end{figure}

The effectiveness of speculative decoding is contingent on the accuracy of the initial predictions made by the draft models, which is critical for the overall decoding speed. Although more complex draft models may provide higher prediction accuracy, they can also lead to increased inference overhead and latency. Research efforts \cite{miao2024specinfer, liu2023online, monea2023pass, spector2023accelerating, zhou2024distillspec, zhang2023draft, hooper2024speed} have primarily investigated the use of independent draft models to enhance latency and throughput. In contrast, recent discussions \cite{cai2024medusa, li2024eagle, bhendawade2024speculative, ankner2024hydra, zhang2024recurrent, zeng2024chimera, du2024glide} have illuminated the benefits of integrated speculators. These integrated approaches are noted for their lightweight architectural design and ease of deployment, offering promising directions for future advancements in speculative decoding.


Figure \ref{fig:intro} shows the Medusa \cite{cai2024medusa} solution, which uses lightweight heads for speculation. It has multiple heads that take inputs from the last transformer block's hidden states, with each layer predicting one token. To address the low hit rate from independent layer speculation, EAGLE \cite{li2024eagle} uses a target model's decoder layer as a draft model to predict tokens iteratively. It combines shifted input embeddings and the last transformer block's hidden states, reducing randomness. However, EAGLE \cite{li2024eagle} encounters challenges, primarily the suboptimal balance between speculative gains and computational expenses when employing the target model's decoder layer. For example, a 5-head sampling necessitates running the decoder layer an additional five times.

To address these challenges, we revisit our proposed \clover framework. \clover is designed specifically for real-time serving scenarios with large inference batch sizes, where traditional speculative decoding frequently encounters computational constraints, resulting in performance degradation. \clover, an RNN architecture with minimal computational requirements, has demonstrated inference speed improvements even with models exceeding 150 billion parameters and batch sizes of 48.

We introduce \Sys, an enhanced version of \clover. The key advancements in \Sys include the \texttt{Information Extraction Order} (Section \ref{sec:pre-set}), the \texttt{Attention Decoder Output Projector} (Section \ref{sec:ouput-proj}), and the \texttt{Augmenting Block} (Section \ref{sec:ab}). These enhancements enable speculators to leverage more sequential knowledge, thereby improving accuracy. Additionally, knowledge distillation (Section \ref{sec:loss-func}) further enhances model training performance.

Tests on Vicuan 7B and LLaMA3-Instruct 8B reveal that \Sys boosts throughput by up to 3.00x over standard decoding and 1.18x-
1.65x over \clover. Despite its RNN architecture, \Sys also delivers a maximum {7.7\%} speculative tokens per step and a maximum {9.3\%} faster speed increase on speculative heads compared to EAGLE.
In summary, our key contributions are:
\begin{itemize}
\item We introduce \Sys, an advanced version of the \clover framework, featuring upgraded model structures and the incorporation of knowledge distillation.
\item Comprehensive evaluations on Vicuan 7B and LLaMA3-Instruct 8B demonstrate that \Sys surpasses the efficiency of \clover and even outperforms EAGLE.

\end{itemize}

\begin{figure}[th]
\centering
\includegraphics[width=0.90\linewidth]{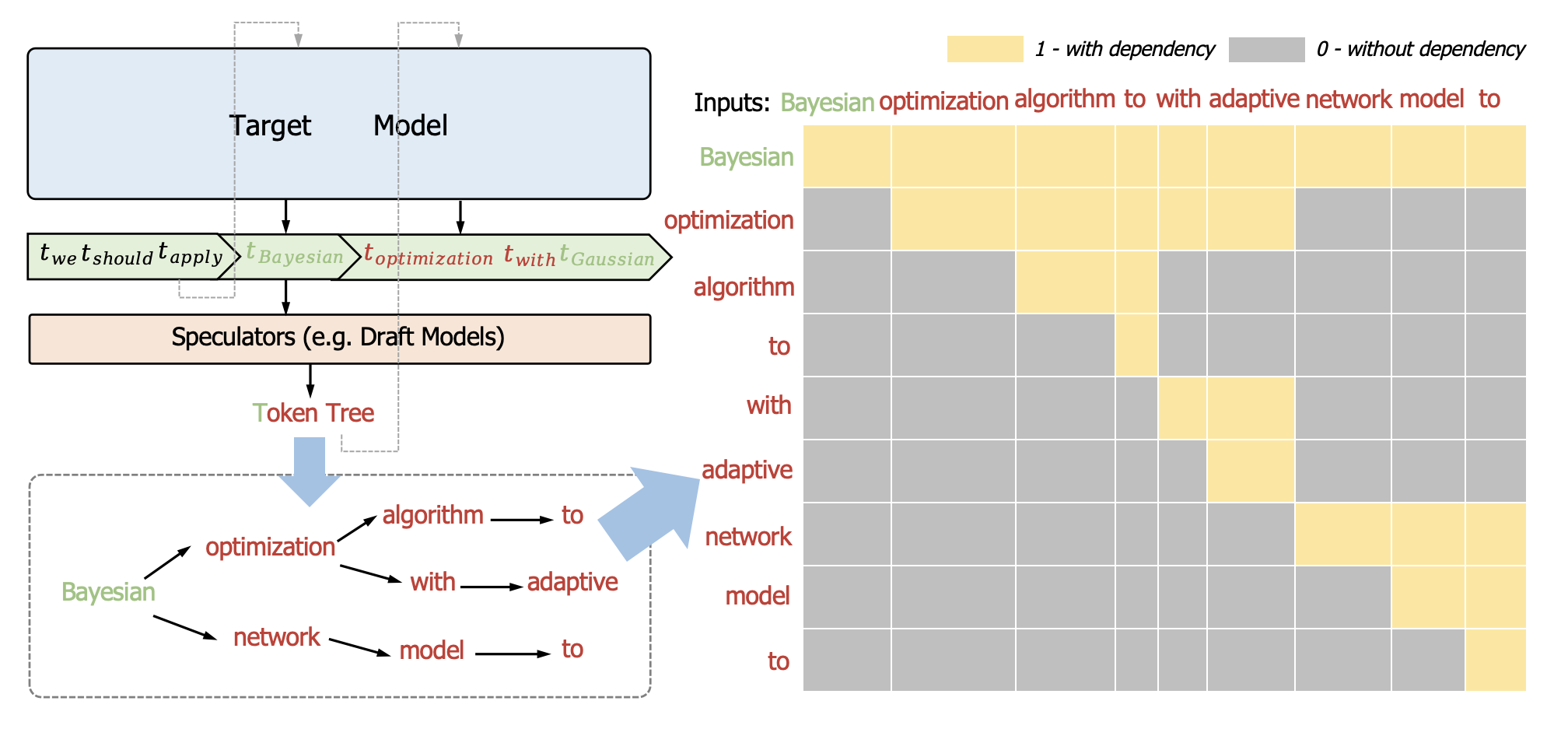}
\caption{A demonstration of Speculative Decoding and Tree Attention. Multiple speculations are merged by prefix matching to form a tree, and its topology dependency is represented in a 2-D matrix as the casual mask in Attention computation.}
\label{fig:tr_aten}
\end{figure}

\section{Background}
\subsection{Speculative Decoding and Tree Attention}

Speculative decoding \cite{pmlr-v202-leviathan23a,chen2023accelerating} represents a sophisticated technique aimed at expediting the inference process of large language models (LLMs) through the enhanced utilization of hardware computational resources.
This method differentiates itself from conventional auto-regressive decoding by concurrently calculating and generating multiple tokens within each iteration.
At the core of speculative decoding resides a speculator component, typically a lightweight model often referred to as the draft model, tasked with predicting multiple subsequent candidate tokens (commonly structured as a tree).
In the context of speculative decoding, the principal LLM, known as the target LLM, ingests all candidate tokens concurrently.
This critical process is designated as the \textit{verification phase}, during which the target LLM meticulously filters out any incorrect tokens from the set of speculative predictions. 
Consequently, speculative inference generates equivalent outputs with a reduced number of decoding steps, thereby significantly enhancing latency efficiency.

Tree Attention \cite{miao2024specinfer} is utilized to calculate attention scores for tree-structured candidate tokens in parallel.
By applying prefix matching to various speculated sequences, the speculation results are organized into a \textit{Token Tree}, which is represented as a 2-D matrix (Figure \ref{fig:tr_aten}).
It is important to note that the attention block is the only component within the modern LLM architecture that requires knowledge of sequential dependency.
The scoring of tree-structured tokens is a relatively straightforward task and can be achieved by configuring the attention's Causal-Mask to align with the topological matrix.
Tree Attention facilitates the integration of multiple speculations with minimal computational overhead, a feature widely implemented in many speculative decoding systems such as \cite{he2024rest, yun2021spectr, xu2023llmcad}.



\subsection{Clover Decoding}
\clover, a lightweight speculative sampling method to address large batch sizes, introduces three incremental components to leverage sequential knowledge: 
\texttt{Regressive Connection}, \texttt{Attention Decoder} and \texttt{Augmenting Block}. 
The Regressive Connection enables sequential dependency from preceding speculated tokens to be considered when a speculator generates the next token.
The Attention Decoder is the factual regressive block in \clover, combining the hidden states from the last transformer block and previously speculated token, merging sequential knowledge between pre-speculated tokens and the entire input sentence.
The Augmentation Block is an additional transformer or self-attention block appended to the target model and is used to enhance sequence features to improve speculator accuracy.

\subsection{EAGLE Decoding}
EAGLE (Extrapolation Algorithm for Greater Language-model Efficiency) \cite{li2024eagle}, a state-of-the-art speculative sampling method, is grounded in two key observations: first, autoregression at the feature level is simpler than at the token level, and second, feature sequences exhibit more regularity compared to token sequences.
By autoregressively processing features and then deriving tokens using the LM head of the original LLM, EAGLE achieves better performance, as evidenced by a higher speedup ratio.
EAGLE incorporates a single transformer decoder layer to the target LLM, ensuring easy deployment in production environments.
Experimental evaluations on various models (Vicuna and LLaMA2-chat series) and tasks (multi-turn dialogue, code generation, mathematical reasoning, instruction following) demonstrate that EAGLE significantly enhances generation speed while maintaining output quality. 
EAGLE's innovative approach of autoregressively processing features and incorporating tokens from one time step ahead effectively mitigates sampling uncertainty, resulting in a substantial acceleration effect.

\section{Clover-2}

\begin{figure}[h]
\centering
\includegraphics[width=.78\linewidth]{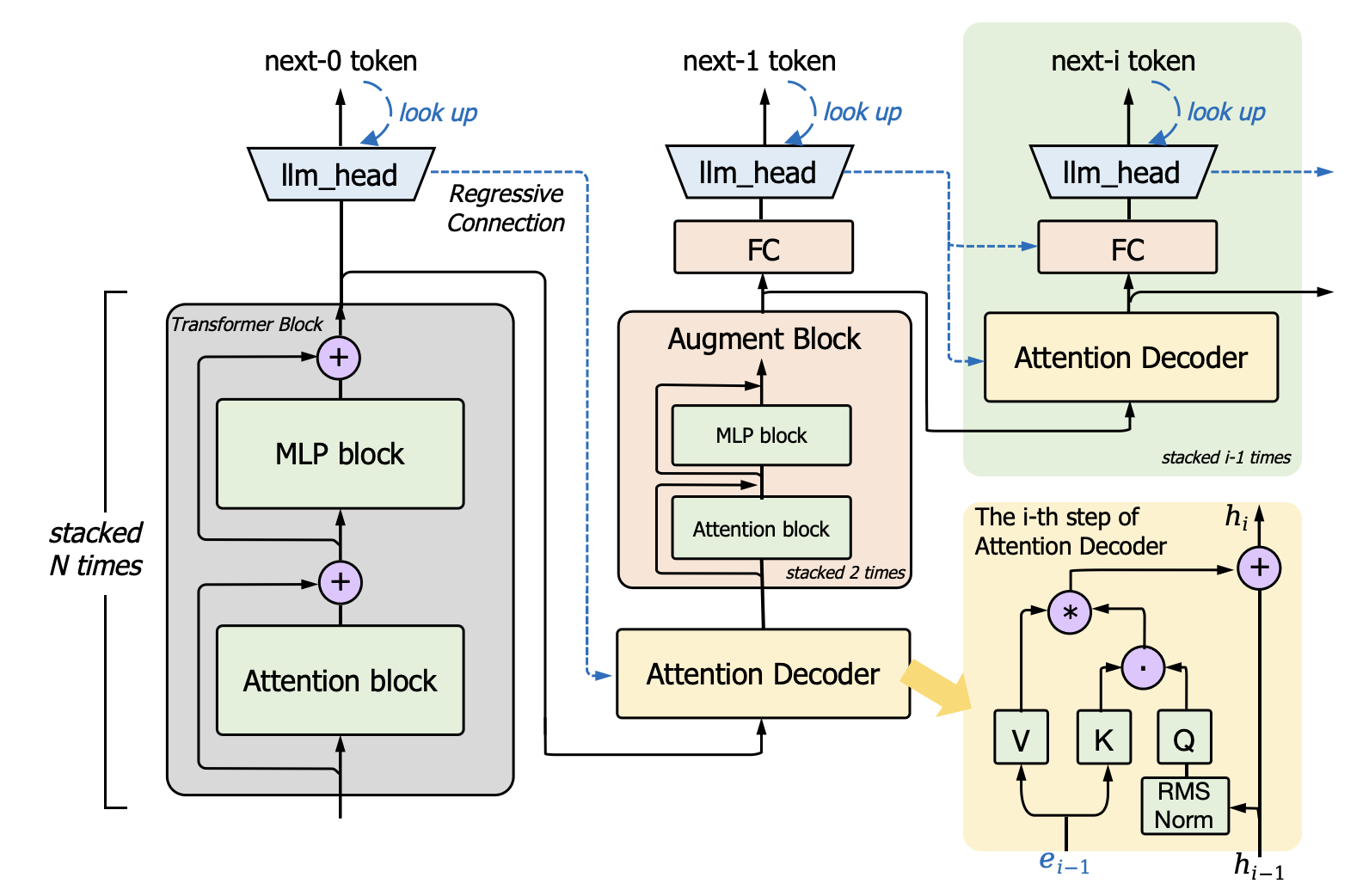}
\caption{Detailed architecture design of \Sys.}
\label{fig:seqar_full}
\end{figure}

Figure \ref{fig:seqar_full} illustrates how \Sys is seamlessly integrated into existing LLMs as the speculator. Like \clover, \Sys incorporates three key functional modules:
\texttt{Regressive Connection}
\texttt{Attention Decoder}
\texttt{Augmenting Block}.
However, there are four notable differences:
(1) The initial head information extraction is predefined. \Sys employs an independent Attention Decoder prior to the Augmenting Block to pre-integrate the hidden states and the output token information of the LLM;
(2) An output projector replaces the ResBlock of Medusa with a fully connected layer, the input to this fully connected layer (FC) encompasses the hidden states and previous token embeddings;
(3) \Sys utilizes a more sophisticated Augment Block to enhance model performance;
(4) \Sys adopts a knowledge distillation strategy, learning not only the classification output of the LLM but also the hidden states of the LLM output.

\subsection{Information extraction order}
\label{sec:pre-set}
The Augmenting Block serves as an excellent sequence information extractor. However, in Clover, the input to the Augmenting Block lacks the information from the last token output by the LLM, potentially undermining its effectiveness. To address this, we introduced an Attention Decoder prior to the Augmenting Block to pre-summarize the hidden states and token information. Consequently, the output projection of the first head bypasses the Attention Decoder, instead directly connecting to a fully connected layer.

\subsection{Attention Decoder output projector}
\label{sec:ouput-proj}
In \Sys, the Attention Decoder output projector, previously a Medusa \cite{cai2024medusa} ResBlock, is replaced with a fully connected layer. This layer accounts for both the hidden states of the Attention Decoder and the token embeddings, thereby mitigating confusion caused by the inherent uncertainty of the hidden states.

A minor adjustment to the Attention Decoder involves the removal of the SiLU activation function. Experiments have indicated that this modification does not result in performance improvements, and it is also deemed anomalous for the residual to only accumulate positive values.

The pseudocode for the Attention Decoder will be presented in Appendix \ref{sec:attentiondecoder}.

\subsection{Augmenting Block} 
\label{sec:ab}
To condense the information from the preceding sequence into a hidden state, \Sys appends n additional transformer blocks following the first Attention Decoder, thereby augmenting features from the entire input sentence. Incorporating such a comprehensive layer incurs a minimal computational overhead (e.g. approximately $1/N_{layer}$ of inference time), while the accuracy gains from the augmenting block far outweigh the time it consumes. The more layers a model possesses, the smaller the proportion of computational consumption becomes.

In EAGLE \cite{li2024eagle}, employing an attention decoder layer as the draft model necessitates running an additional number of head layers of attention decoder layers for each decode process. \Sys utilizes a lightweight Attention Decoder, with a computational load approximately 2.5 times lighter than a single layer of EAGLE, enabling the use of a more computationally intensive Augmenting Block. Such an approach is not feasible in EAGLE\cite{li2024eagle}, where any additional operations incur costs that must be multiplied by the number of heads. \Sys adopts the simplest method, increasing the number of decoder layers in the Augmenting Block to 2.

\subsection{Knowledge distillation} 
\label{sec:loss-func}
During the comparative training between \clover and Eagle\cite{li2024eagle}, \clover displayed severe overfitting. Various strategies were tested without any improvement. Eventually, we observed Eagle's regression loss, which was only mentioned in the paper for auxiliary intermediate result learning. Through analysis and experimentation, we discovered that regression loss enables the draft model to focus not only on the probability of output tokens but also to more closely align with the distribution of the LLM. This represents a more profound knowledge distillation strategy, effectively suppressing overfitting and enhancing model performance. Regression loss calculates the L1 loss using the LLM's output hidden states (after normalization) and the hidden states (after normalization) output by the draft model. In \Sys, we refer to it as regularization loss. Consequently, our loss function was updated as follows:

\begin{equation}
L_{reg}{i}=Smooth\_L1(LLM\ hidden\ states_{i+1},\ Draft\ hidden\ states_{i}).
\end{equation}
\begin{equation}
L_{cls}{i}=CrossEntropy(LLM\ prob_{i+1},\ Draft\ prob_{i}).
\end{equation}
\begin{equation}
L=\sum_{i=0}^{n-1} (L_{cls}{i} \  + \  w\_reg \ * \ L_{reg}{i}) \ * \ decay\_coefficient ^ i.
\end{equation}
, where $n$ denotes the number of draft model heads, which is 5 in \Sys. In the optimal practices of \Sys, $w\_reg$ is set to 10.0 and $decay\_coefficient$ is set to 0.7.

\subsection{Other Details}

Firstly, in \clover, a layer normalization is incorporated prior to the Attention Decoder, whereas in \Sys, a layer normalization is introduced before the second Attention Decoder. Additionally, a layer normalization is applied before the llm\_head, mirroring the configuration of the LLM. Similar to \clover, token embeddings are derived from the transposed matrix of the llm head weight.

Secondly, during the design process of \Sys, it was observed that the actual training data is not necessarily SFT data for LLM, and even SFT data exhibits distribution differences compared to data directly decoded by LLM. To address this, we devised a sample mask strategy. Based on the token probability output by the model, we select top\_k, top\_p and compare it with the ID of the next token. If it falls within the set, the token is retained. Concurrently, different heads will be connected in series with the mask of the previous token according to the decode method. For instance, if head 1 is masked, subsequent heads will also be masked. In experiments with Llama-8B, no gains were observed. We are currently analyzing the specific reasons. Preliminary analysis suggests that the draft model learns relatively simple aspects, leaving complexity unlearned. Adding these samples is akin to introducing noise, which can prevent overfitting.

Lastly, we also designed a compressed tree mask structure, which is an additional design. This section will be included in Appendix \ref{sec:tree_mask}.

\section{Evaluation}

\subsection{Experiment Settings}

\paragraph{Models and baselines} Both the EAGLE and \Sys approaches are employed on the \texttt{Vicuna 7B v1.5} \cite{vicuna} and \texttt{LLaMA3-Instruct 8B}  models \cite{llama3} with the number of speculative head is 5.
To ensure the fairness of the comparison, the inference engine, tree construction and tree sampling algorithm of EAGLE
are used for all scenarios. We also evaluate auto-regressive decoding under the same circumstances. 

\paragraph{Dataset and Metrics} We employ the SharedGPT dataset, containing 68,000 dialogue iterations, to train both EAGLE and \Sys.
We then evaluate inference performance on Spec-Bench\cite{xia2024unlocking}, 
which includes data from MT-bench \cite{zheng2024judging}, WMT14 DE-EN (WMT14) \cite{bojar2014findings}, CNN/Daily Mail (CNN/DM) \cite{nallapati2016abstractive}, Natural Questions (NQ) \cite{kwiatkowski2019natural}, GSM8K \cite{cobbe2021training}, and DPR \cite{karpukhin2020dense}, representing the tasks of multi-turn conversation, translation, summarization, question answering, mathematical reasoning, and retrieval-augmented generation, respectively.
We choose extra generated tokens (i.e. \texttt{tokens/step}) and throughput (i.e. \texttt{tokens/second}) as our main metrics, followed by prior speculative decoding works.

\paragraph{Training} Both models are trained with all weights frozen in the target model. 
For EAGLE, the initial weight settings correspond to the configuration given in \cite{li2024eagle}.
While for \Sys, the initial weight settings correspond to the configuration given in Appendix \ref{sec:init}.
We train \Sys for 20 epochs (about 4000 steps per epoch), with  $(\beta_1=0.9,\beta_2=0.95)$ for the AdamW optimizer.
The learning rate\footnote{Linear decay is applied to the learning rate.} is set to 1e-3 with linear schedule(warmup-steps=1000, final-min-lr=5e-4).


\subsection{End-to-end Results}
\begin{figure}[th]
    \centering
    \includegraphics[width=0.85\linewidth]{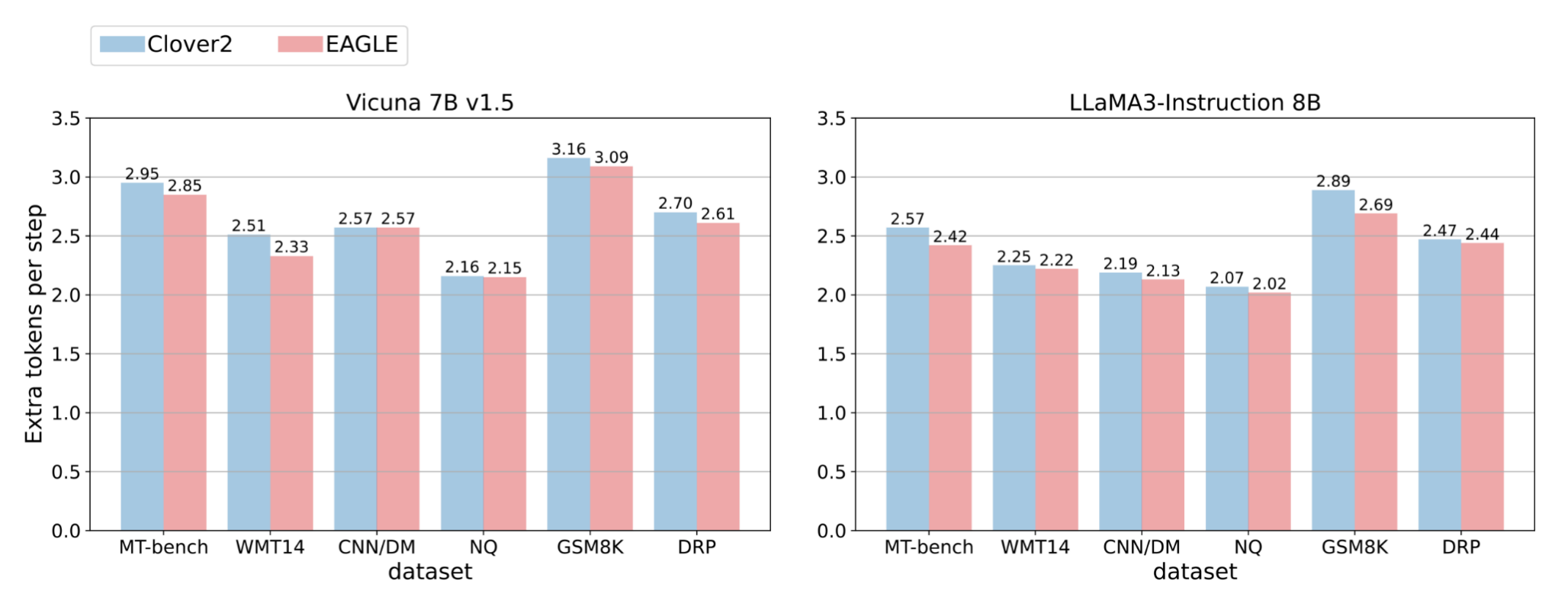}
    \caption{Number of extra generated tokens (excluding the first one) per step on various tasks.}
    \label{fig:tokperstep}
\end{figure}

\begin{table}[t]
    \centering
    \small
    \begin{tabular}{|P{1cm}|P{1.3cm}|P{1.4cm}P{1.0cm}P{1.0cm}P{1.0cm}P{1.0cm}P{1.0cm}|}
        \hline
        \multirow{2}{*}{Model} & \multirow{2}{*}{Approach}  & \multicolumn{6}{c|}{Tokens/second and Speedup rate over Vanilla Decoding} \\
        &&MT-bench&WMT14&CNN/DM&NQ&GSM8K&DPR\\\hline\hline 
\multicolumn{8}{|c|}{Temperature = 0} \\
\hline
\rule{0pt}{2.0ex}
        \multirow{5}{*}{V 7B} & \multirow{2}{*}{\centering \Sys}  &\textbf{145.6}&\textbf{111.0}&\textbf{121.9}&\textbf{115.0}&\textbf{149.2}&\textbf{107.6}\\
        &&\textbf{3.00x}&\textbf{2.43x}&\textbf{2.55x}&\textbf{2.40x}&\textbf{3.09x}&\textbf{2.44x}\\\cline{2-8}
        \rule{0pt}{2.0ex}
        & \multirow{2}{*}{\centering EAGLE}&142.6&106.8&120.6&107.4&141.9&102.8\\
        &&2.94x&2.34x&2.52x&2.24x&2.94x&2.34x\\
        \cline{1-8}


\rule{0pt}{2.0ex}
        \multirow{5}{*}{L 8B} & \multirow{2}{*}{\centering \Sys}  &\textbf{121.0}&\textbf{108.6}&\textbf{99.5}&\textbf{103.2}&\textbf{132.0}&\textbf{97.4}\\
        &&\textbf{2.47x}&\textbf{2.24x}&\textbf{2.09x}&\textbf{2.08x}&\textbf{2.65x}&\textbf{2.14x}\\\cline{2-8}
        \rule{0pt}{2.0ex}
        & \multirow{2}{*}{\centering EAGLE}&113.6&106.7&96.0&100.1&123.4&96.5\\
        &&2.32x&2.20x&2.02x&2.02x&2.47x&2.12x\\
        \hline


\multicolumn{8}{|c|}{Temperature = 1} \\
\hline
\rule{0pt}{2.0ex}
        \multirow{5}{*}{V 7B} & \multirow{2}{*}{\centering \Sys}  &\textbf{112.9}&\textbf{90.0}&\textbf{98.0}&\textbf{93.2}&\textbf{116.9}&\textbf{87.8}\\
        &&\textbf{2.25x}&\textbf{1.95x}&\textbf{2.03x}&\textbf{1.91x}&\textbf{2.39x}&\textbf{1.98x}\\\cline{2-8}
        \rule{0pt}{2.0ex}
        & \multirow{2}{*}{\centering EAGLE}&109.9&82.3&95.1&91.4&113.6&87.7\\
        &&2.19x&1.79x&1.97x&1.87x&2.32x&1.98x\\
        \cline{1-8}


\rule{0pt}{2.0ex}
        \multirow{5}{*}{L 8B} & \multirow{2}{*}{\centering \Sys}  &\textbf{98.5}&\textbf{87.6}&\textbf{81.3}&\textbf{81.6}&\textbf{105.5}&\textbf{81.6}\\
        &&\textbf{2.16x}&\textbf{1.93x}&\textbf{1.80x}&\textbf{1.77x}&\textbf{2.28x}&\textbf{1.90x}\\\cline{2-8}
        \rule{0pt}{2.0ex}
        & \multirow{2}{*}{\centering EAGLE}&91.3&84.7&78.1&78.0&97.9&78.9\\
        &&2.00x&1.87x&1.73x&1.69x&2.11x&1.84x\\
        \hline
    \end{tabular}
    \vspace{10pt}
    \caption{End-to-end throughput on Vicuan 7B v1.5 (V 7B) and LLaMA3-Instruction 8B (L 8B) with different decoding methods on six tasks. Temperature value of 0 represents greedy decoding for the target LLM, while Temperature value of 1 represents non-greedy decoding.}
    \vspace{-10pt}
    \label{tab:e2e}
\end{table}

We evaluate the end-to-end performance at different target LLMs and tasks.
Figure \ref{fig:tokperstep} illustrates the average number of tokens generated per step for \Sys and EAGLE methods on different tasks with greedy decoding. 
Note that the value on the vertical axis is the \textbf{extra} tokens per step, excluding the actual token generated by target LLM, which more accurately reflects the performance of the speculator. 
With the support of the model structure performance, \Sys generates more tokens per step as EAGLE across all tasks. For model \texttt{Vicuna 7B v1.5}, a maximum of {7.7\%} and an average of {2.9\%} improvement; for model \texttt{LLaMA3-Instruction 8B}, a maximum of {7.4\%} and an average of {3.6\%} improvement.

Table \ref{tab:e2e} displays the end-to-end throughput (i.e., \texttt{tokens/second}) and the speedup rate relative to Vanilla Decoding.
The results indicate that both methods achieve speedup across all tasks when compared to Vanilla Decoding.
In the case of temperature being set to 0,
\texttt{Vicuna 7B v1.5} model shows a maximum improvement of {7.1\%} and an average improvement of {4.0\%}; the \texttt{LLaMA3-Instruction 8B} model exhibits a maximum improvement of {7.0\%} and an average improvement of {3.8\%}. 
When the temperature is set to 1, 
the \texttt{Vicuna 7B v1.5} model demonstrates a maximum improvement of {9.3\%} and an average improvement of {3.3\%}, while the \texttt{LLaMA3-Instruction 8B} model presents a maximum improvement of {7.9\%} and an average improvement of {5.3\%}.


It should be emphasized that the above inference framework and sampling method use the same approach as Eagle.
The framework is not an efficient implementation, the provided data is for reference purposes only.
In theory, the more efficient the framework, the greater the benefits of \Sys, because \Sys has lower computational requirements and subsequent heads do not need to construct complex attention-related parameters.

\subsection{Ablation Study}

In the ablation study, we gradually add modules according to the experimental timeline to measure the effectiveness of each module compared to \clover. The main metric is the extra generated tokens (i.e., \texttt{tokens/step}). Clover2 has an average improvement of about {30\%} compared to Clover, with relevant data presented in Table \ref{tab:AS}. The benefits brought by each module are as follows:
\paragraph{Knowledge distillation} In the comparative experiment between \clover and EAGLE, severe overfitting was observed in \clover. To address this issue, we introduced a regularization loss based on knowledge distillation, which contributed to a {9\%} performance improvement. The main improvement comes from the later epochs, which continuously enhance the metrics.
\paragraph{Information extraction order} Pre-setting the information aggregation of the first head allows for full utilization of the Augmenting Block's sequence extraction capabilities, effectively raising the performance ceiling of the draft model. This optimization resulted in a {7\%} improvement.
\paragraph{Attention Decoder output projector} Modifying the output projector significantly improved the hit rate of the latter heads, contributing to a {5\%} gain.
\paragraph{Augmenting Block} Enhancing the number of layers within the Augmenting Block effectively strengthens the sequence information aggregation capability, providing a {9\%} overall benefit.

\begin{table}[t]
    \centering
    \small
    \begin{tabular}{|P{1cm}|P{1.3cm}|P{1.4cm}P{1.0cm}P{1.0cm}P{1.0cm}P{1.0cm}P{1.0cm}|}
        \hline
        \multirow{2}{*}{ } & \multirow{2}{*}{Approach}  & \multicolumn{6}{c|}{Tokens/step and Speedup rate over Vanilla Decoding} \\
        &&MT-bench&WMT14&CNN/DM&NQ&GSM8K&DPR\\\hline
\rule{0pt}{2.0ex}
        \multirow{4}{*}{T = 0} & \multirow{2}{*}{\centering \Sys}  &\textbf{2.95}&\textbf{2.51}&\textbf{2.57}&\textbf{2.16}&\textbf{3.16}&\textbf{2.70}\\
        &&\textbf{3.00x}&\textbf{2.43x}&\textbf{2.55x}&\textbf{2.40x}&\textbf{3.09x}&\textbf{2.44x}\\\cline{2-8}
        \rule{0pt}{2.0ex}
        & \multirow{2}{*}{\centering \clover}&2.46&1.78&1.48&1.67&2.52&1.78\\
        &&2.27x&1.70x&1.54x&1.71x&2.15x&1.66x\\
        \cline{1-8}


\rule{0pt}{2.0ex}
        \multirow{4}{*}{T = 1} & \multirow{2}{*}{\centering \Sys}  &\textbf{2.50}&\textbf{2.15}&\textbf{2.22}&\textbf{1.82}&\textbf{2.79}&\textbf{2.24}\\
        &&\textbf{2.25x}&\textbf{1.95x}&\textbf{2.03x}&\textbf{1.91x}&\textbf{2.39x}&\textbf{1.98x}\\\cline{2-8}
        \rule{0pt}{2.0ex}
        & \multirow{2}{*}{\centering \clover}&2.03&1.60&1.33&1.53&2.16&1.62\\
        &&1.88x&1.59x&1.44x&1.61x&1.94x&1.57x\\
        \hline

    \end{tabular}
    \vspace{10pt}
    \caption{Ablation study on Vicuan 7B v1.5 with different decoding methods on six tasks, where \texttt{T} in the head means temperature.}
    \vspace{-10pt}
    \label{tab:AS}
\end{table}

\section{Related Works}
Since the introduction of speculative decoding for LLMs as outlined in \cite{pmlr-v202-leviathan23a,chen2023accelerating}, numerous optimization techniques have been developed.
The concept of tree attention, as explored in \cite{miao2024specinfer}, has been widely implemented for the efficient verification of multiple speculations in a single step. 
Initial research efforts \cite{NEURIPS2023_7b97adea,liu2023online,monea2023pass,spector2023accelerating,zhou2024distillspec,zhang2023draft,hooper2024speed,chen2024cascade} concentrated on enhancing independent draft models. In contrast, later studies \cite{yang2023inference,he2024rest,fu2024break} focused on draft model architectures that do not require additional training.
More contemporary research has delved into the potential advantages of regressive speculators.
Zhang et al. \cite{zhang2024recurrent} employ a Multilayer Perceptron (MLP) layer as a regression block, Hydra \cite{ankner2024hydra} and ReDrafter \cite{zhang2024recurrent} introduce a regressive component based on a Recurrent Neural Network (RNN), Eagle \cite{li2024eagle} incorporates a transformer decoder layer for speculation, Chimera \cite{zeng2024chimera} proposes the utilization of a Trigram Encoder and a Full Context Encoder as sophisticated regressive speculation mechanisms.
The main difference in \Sys is the use of the Attention Decoder and Augment Block to capture sequential context information, followed by an RNN architecture to output multiple candidate tokens.

\section{Conclusion}
We present an upgraded version of \clover, named \Sys, which incorporates four enhancement points(Section \ref{sec:pre-set}, \ref{sec:ouput-proj}, \ref{sec:ab}, \ref{sec:loss-func}).
In tests conducted against the original \clover and the current state-of-the-art EAGLE, \Sys not only significantly boosts the performance of \clover but also surpasses EAGLE in terms of hit rate and speed.
Relative to \clover, \Sys achieves at least {19\%} increase in speculative tokens per step and an {18\%} improvement in speed.
When compared to EAGLE, a state-of-the-art method, \Sys shows a maximum {7.7\%} more speculative tokens per step and a maximum {9.3\%} faster speed. These results demonstrate the effectiveness of the implemented improvements.



\bibliographystyle{plain}
\bibliography{egbib}

\appendix
\newpage
\section{Appendix}

\subsection{Compressed tree mask}
\label{sec:tree_mask}
\begin{figure}[htb]
    \vspace{-10pt}
    \centering
    \includegraphics[width=0.82\linewidth]{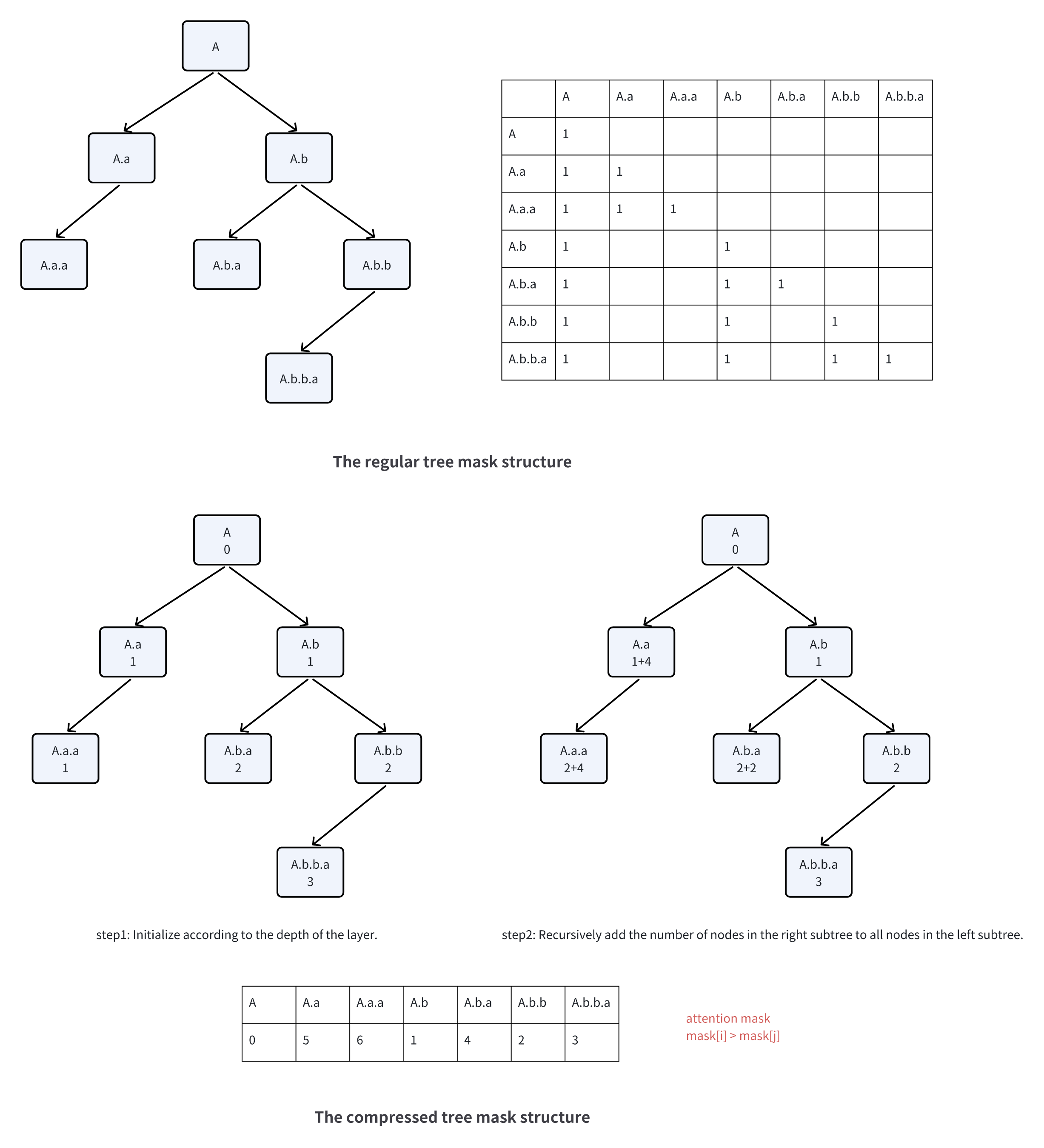}
    \caption{The difference between compressed and regular tree mask structure}
    \label{fig:tree_size}
\end{figure}
As shown in Figure \ref{fig:tree_size}, we designed a linear tree mask structure, confirming the mask relationship through numerical comparison.

\subsection{Model parameter initialization methods}
\label{sec:init}
\begin{table}[htb]
\centering
\small
\begin{tabular}{|P{2.5cm}|P{10.5cm}|}
\hline
Parameter & Init Method \\
\hline
\textbf{embedding lookup} & lm\_head weight Matrix transpose \\
\hline
\textbf{Augmenting Block} & last decoder layer of base model  \\
\hline
\textbf{head 0 FC} & eyes and uniform(b=0.01)  \\
\hline
\textbf{Attention Decoder 1st/2nd} & q/k with eyes and uniform(b=0.01), v with uniform(b=0.01), bias with zero \\
\hline
\textbf{head 1-n FC} & eyes with uniform(b=0.01) for hidden state part,  uniform(b=0.01) for embeding part \\
\hline
\textbf{norm 1st/2nd} & base model norm \\\hline
\end{tabular}
\vspace{10pt}
\caption{\Sys Model parameter initialization methods.}
\vspace{-10pt}
\end{table}

\newpage

\subsection{The pseudocode of Attention Decoder}
\label{sec:attentiondecoder}
\begin{python}

class AttentionDecoder(nn.Module):

    def __init__(self, hidden_size, head_size, rms_norm_eps):
        super().__init__()
        self.head_size = head_size
        self.head_dim = hidden_size // head_size
        assert hidden_size 
        self.layernorm = LlamaRMSNorm(hidden_size, rms_norm_eps)
        self.q = nn.Linear(hidden_size, hidden_size)
        self.k = nn.Linear(hidden_size, hidden_size)
        self.v = nn.Linear(hidden_size, hidden_size)

    def forward(self, x, y):
        res = x
        x = self.input_layernorm(x)
        x_q = self.q(x)
        y_k = self.k(y).view(-1, self.head_dim)
        att = cosine_similarity(x_q.view(-1, self.head_dim), y_k)
        att = att.view(-1, self.head_size, 1)
        v =  self.v(y).view(-1, self.head_size, self.head_dim)
        v = v * att
        v = v.view(x_q.size())
        return res + v

\end{python}


\end{document}